\documentclass{article}
\usepackage{PRIMEarxiv}
\usepackage[utf8]{inputenc} 
\usepackage[T1]{fontenc}    
\usepackage{hyperref}       
\usepackage{url}            
\usepackage{booktabs}       
\usepackage{amsfonts}       
\usepackage{amssymb}        
\usepackage{nicefrac}       
\usepackage{microtype}      
\usepackage{lipsum}
\usepackage{fancyhdr}       
\usepackage{graphicx}       
\usepackage{amsmath}       
\graphicspath{{media/}}     
\usepackage{algorithm}
\usepackage{multirow}
\usepackage{algpseudocode}
\title{GoQuant: Geometric Orthogonal Residual Projection for Multiplier-Reduced Power-of-Two Transformer Quantization
\thanks{
\textbf{In preparation.}}
}

\author{
  Maoyang Xiang \\
  Information Systems Technology and Design \\
  Singapore University of Technology and Design \\
  Singapore\\
  \texttt{maoyang\_xiang@sutd.edu.sg} \\
   \And
  Tao Luo \\
  Institute of High Performance Computing (IHPC) \\
  Agency for Science, Technology and Research (A*STAR) \\
  Singapore\\
  \texttt{luo\_tao@a-star.edu.sg} \\
  \And
  Bo Wang \\
  Information Systems Technology and Design \\
  Singapore University of Technology and Design \\
  Singapore\\
  \texttt{bo\_wang@sutd.edu.sg} \\
}

\begin{document}
\maketitle

\begin{abstract}
The deployment of Large Language Models (LLMs) and Vision Transformers (ViTs) on edge devices is significantly constrained by memory capacity and the critical timing bottlenecks introduced by dense Multiply--Accumulate (MAC) arrays. In the ultra-low-bit regime, logarithmic Power-of-Two (PoT) quantization provides a hardware-efficient alternative by replacing general multiplications in the dominant dot-product computation with bit-shift operations. However, its non-uniform exponential lattice inherently suffers from a \textbf{Low Angular Resolution Regime}, a structural limitation that becomes particularly pronounced below 4-bit precision and can substantially degrade the representation of high-dimensional feature manifolds.

To address this geometric limitation, we propose Geometric Orthogonal Residual Projection Quantization (GoQuant), an algorithm--hardware co-design framework for multiplier-reduced low-bit inference. By formulating quantization as a dual-basis geometric projection, GoQuant constructs a higher-resolution residual lattice while retaining a shift-and-add inner-product structure. Its analytical solver further avoids computationally intensive gradient-based or iterative search procedures. The data-free Geometric-Only (GEO) mode quantizes LLaMA-2-7B in only 0.47 minutes, while the Activation-Refined (REF) mode completes full-model quantization in approximately \textbf{4.4 minutes}.

Extensive evaluations demonstrate the cross-modality applicability, model fidelity, and hardware efficiency of GoQuant. Under W3/A16 quantization, GoQuant achieves a WikiText-2 perplexity of \textbf{6.01} on LLaMA-2-7B, improving upon the 6.49 perplexity of AWQ while avoiding runtime affine reconstruction of the stored PoT operands before the core dot-product computation. It also maintains competitive accuracy across multiple Vision Transformer architectures through flexible weight--activation precision configurations. At the hardware level, standard-cell RTL synthesis at a 28\,nm technology node shows that the proposed parallel shift-and-add inner-product datapath consistently reduces both the energy--delay product and critical-path delay relative to conventional multiplier-based arithmetic units. Under W3/A8, GoQuant achieves a minimum critical-path delay of \textbf{0.32\,ns}, corresponding to a nominal synthesized-unit frequency of approximately \textbf{3.13\,GHz}. Lightweight coefficient scaling remains outside the dominant shift-and-add dot-product core. These results establish GoQuant as an efficient and flexible multiplier-reduced quantization paradigm for deploying large language and vision models on resource-constrained edge platforms.
\end{abstract}


\section{Introduction}

The deployment of large language models (LLMs) and vision transformers (ViTs) on edge devices is frequently constrained by the \textit{multiplier tax}---the critical combinational logic depth and routing complexity associated with hardware multiply--accumulate (MAC) units. General multipliers typically introduce deeper logic paths and greater switching and routing costs than additions or bit-wise operations. To advance high-throughput edge computing, there is therefore strong motivation to reduce multiplication in the dominant matrix dot products. In this context, \textbf{Power-of-Two (PoT)} quantization offers a hardware-oriented alternative in which PoT operands can be applied through parallel shifts and additions, while lightweight coefficient rescaling may remain outside the inner-product core.

\begin{figure}[htbp]
    \centering
    \includegraphics[width=0.55\linewidth]{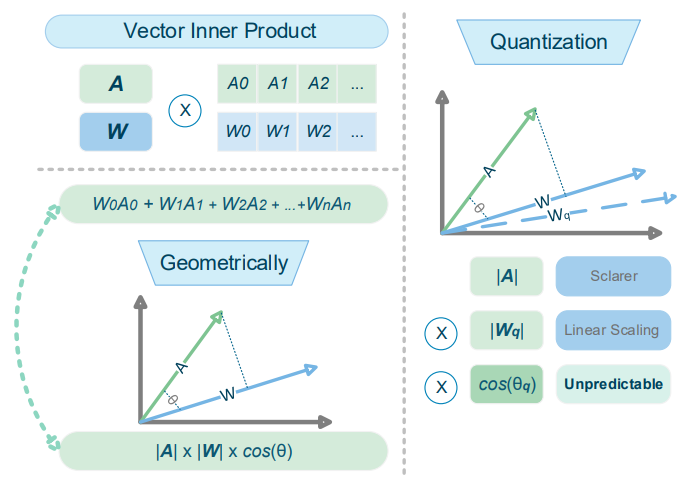}
    \caption{The geometric worldview of quantization. Beyond simple algebraic accumulation (top), the vector inner product is fundamentally a geometric projection (bottom). While magnitude errors (\textit{scale}) are predictable scalar factors that can be easily calibrated, standard quantization introduces unpredictable and uncorrectable angular deviation ($\cos\theta_q$).}
    \label{fig:s1_geometric_view}
\end{figure}

Despite its hardware advantages, low-bit quantization often faces challenges regarding precision. Conventional approaches typically attribute this performance degradation to a scalar distribution mismatch between the discrete lattice and the Gaussian statistics of neural weights. However, this scalar-centric perspective often overlooks the high-dimensional nature of Transformer operations\cite{chen2025geometry}. As illustrated in Fig.~\ref{fig:s1_geometric_view}, the inner product is essentially the projection of a continuous weight vector onto a discrete lattice manifold. In high-dimensional spaces, the fidelity of this projection depends not just on magnitude, but critically on the preservation of the vector's direction, which dictates the alignment of latent features. While current methods effectively manage the scaling factor (\textit{scale}), directional integrity can remain unpredictable, contributing significantly to performance degradation in extreme quantization.

\begin{figure}[htbp]
    \centering
    \includegraphics[width=0.4\linewidth]{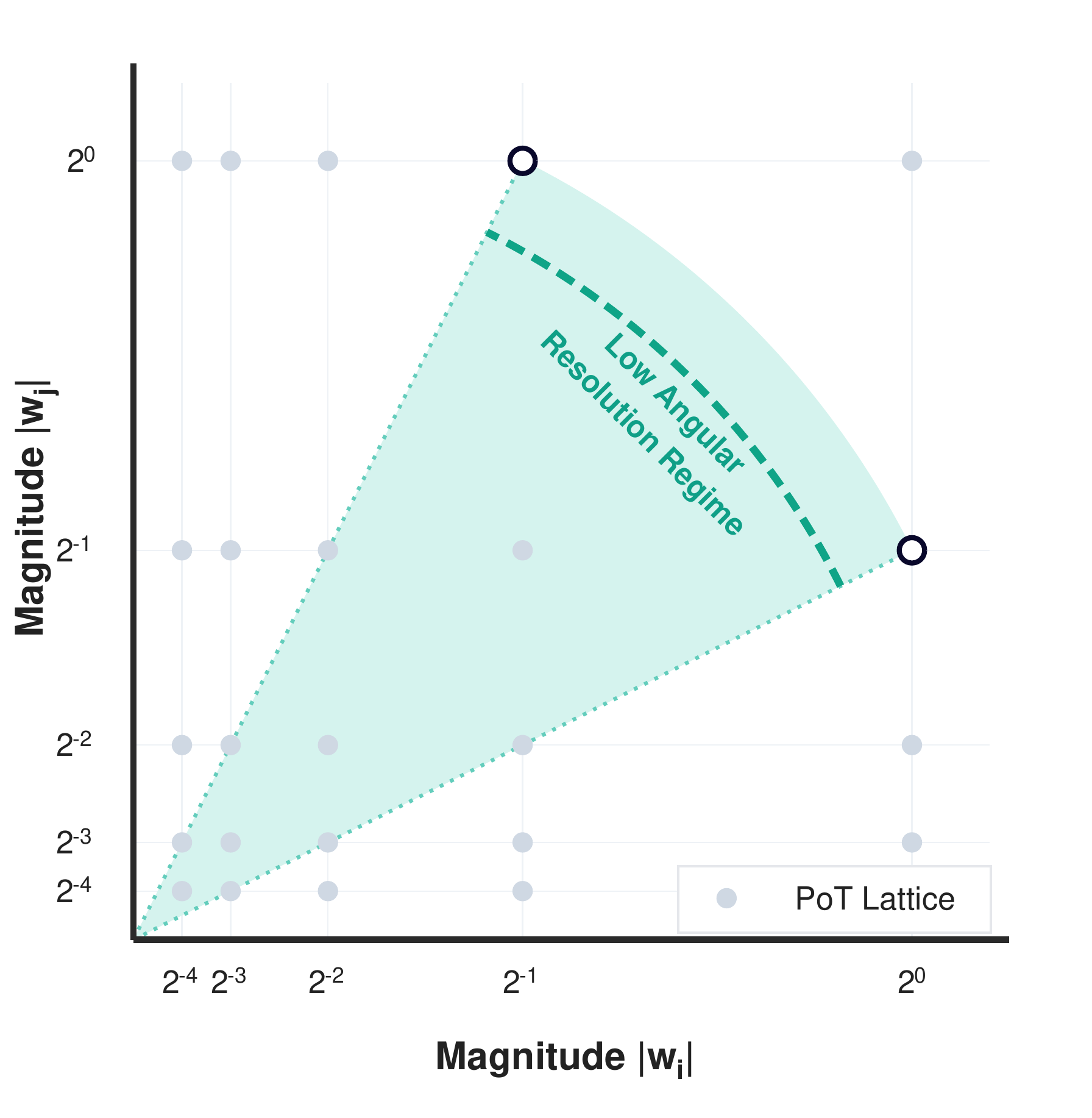}
    \caption{The Power-of-Two (PoT) lattice structure exhibits a non-uniform angular resolution, particularly in high-magnitude regions. As illustrated by the shaded area, the grid density decreases significantly, resulting in a \textbf{Low Angular Resolution Regime}. This structural characteristic introduces notable \textbf{Angular Gaps} between valid lattice directions; these gaps persist unless the exponent spacing itself is refined.}
    \label{fig:s1_anglehole}
\end{figure}

When applying this geometric perspective to the hardware-friendly PoT domain, a key limitation emerges: its inherently non-uniform angular resolution. Unlike linear lattices, the angular distribution of PoT basis vectors is governed by the $\arctan(2^k)$ function. Consequently, it operates within a \textbf{low angular resolution regime}, as shown in Fig.~\ref{fig:s1_anglehole}, which is particularly pronounced around the diagonal regions of the projection space. This structural characteristic creates an \textbf{Angular Gap}, where the quantized vector's orientation can deviate enough to substantially compromise the semantic integrity of the feature manifold.

To address the angular degradation associated with a one-dimensional PoT mapping, we introduce the \textbf{Geometric Orthogonal Residual Projection Quantization (GoQuant)} framework. Rather than mapping a continuous high-dimensional vector onto a single discrete direction ($\mathbf{b}_1$), GoQuant constructs a two-dimensional orthogonal subspace. A secondary basis ($\mathbf{b}_2$) is generated such that $\langle \mathbf{b}_1,\mathbf{b}_2\rangle=0$, allowing it to capture directional information left unresolved by the primary PoT projection. This dual-basis representation mitigates the identified low-angular-resolution behavior while preserving a shift-and-add inner-product implementation.

GoQuant employs a decoupled two-stage pipeline. First, a geometric residual-matching procedure constructs the secondary discrete basis $\mathbf{b}_2$ directly from the primary PoT projection and its orthogonal residual. Because this basis construction is independent of activation data, it admits a deterministic, hardware-bounded $\mathcal{O}(N)$ solution without gradient-based optimization. Second, a joint coefficient solver determines the continuous projection coefficients $c_1$ and $c_2$. The data-free GEO mode solves these coefficients in the Euclidean weight space, whereas the calibration-based REF mode incorporates empirical activation statistics to prioritize output reconstruction along feature directions represented in the calibration set.

Evaluating multiplier-reduced networks solely on algorithmic metrics may not fully reflect their physical deployment constraints. To address the gap between theoretical sparsity and practical silicon implementation, we extend our evaluation to standard-cell RTL synthesis, analyzing the timing implications associated with conventional arithmetic units.

In summary, the main contributions of this work are as follows:

\begin{itemize}
\item We identify and geometrically characterize the \textbf{Low Angular Resolution Regime} inherent to extreme PoT quantization and introduce GoQuant as a geometric solution based on structured orthogonal residual projection. GoQuant deterministically constructs complementary discrete bases through analytical sign inference and a bounded set of hardware-supported exchange patterns, yielding an $\mathcal{O}(N)$ quantization procedure without gradient-based optimization or high-dimensional iterative search.

\item By providing an analytical alternative to computationally intensive backpropagation and iterative optimization, GoQuant substantially accelerates the deployment pipeline. On a single workstation, the data-free GEO mode completes full-model quantization in only \textbf{0.47 minutes} for LLaMA-2-7B, while the activation-refined REF mode requires approximately \textbf{4.4 minutes}. For LLaMA-2-13B, the corresponding quantization times are less than one minute for GEO and approximately \textbf{10.4--10.5 minutes} for REF, enabling practical on-site adaptation of low-bit edge models.

\item GoQuant demonstrates that strict hardware constraints can still preserve competitive representational fidelity across modalities. Using a reconstruction-free, shift-and-add inner-product datapath, it achieves a WikiText-2 perplexity of \textbf{5.10} on LLaMA-2-13B under \textbf{W4/A16} quantization and a Top-1 accuracy of \textbf{79.84\%} on ViT-Base under \textbf{W4/A4}. In the more challenging W3/A16 regime, GoQuant further improves LLaMA-2-7B perplexity from 6.49 with AWQ to \textbf{6.01}.

\item We evaluate the hardware efficiency of the proposed architecture through standard-cell RTL synthesis at a 28\,nm technology node. By replacing dense multiplier trees with a parallel shift-and-add datapath, GoQuant consistently reduces both the energy--delay product and critical-path delay relative to conventional multiplier-based arithmetic units. Under W3/A8, the design achieves a minimum critical-path delay of \textbf{0.32\,ns}, corresponding to a nominal synthesized-unit frequency of approximately \textbf{3.13\,GHz}, while attaining an EDP of \textbf{0.32\,pJ$\cdot$ns}. The localized 8-lane exchange architecture further limits each runtime selection network to a compact 4-to-1 multiplexer, reducing multiplexer depth, cross-lane wiring, and routing pressure. Each micro-block stores a 2-bit exchange-pattern index and four pairwise sign bits, corresponding to 6 bits per 8 weights, or 0.75 bit per weight, in addition to the PoT codes and block coefficients.
\end{itemize}

\section{Related Work}

The deployment of large-scale foundation models on resource-constrained edge platforms has driven extensive research into neural network quantization \cite{xiao2023smoothquant,frantar2022gptq}. While significant progress has been achieved in compressing model weights, simultaneously maintaining ultra-low bit-width accuracy and mitigating computationally heavy hardware operations remains a notable challenge. In this section, we review the prior literature across three pivotal domains. First, we examine the evolution of Post-Training Quantization (PTQ) and the hardware implications of reconstruction-heavy paradigms. Second, we analyze recent advancements in high-dimensional rotational quantization, highlighting the trade-off between geometric angular resolution and on-chip computational overhead. Finally, we discuss the trajectory of multiplier-reduced architectures, examining the representational limitations of scalar logarithmic quantization and demonstrating how our proposed framework aims to address these challenges.

\subsection{Post-Training Quantization and the Reconstruction Bottleneck}
Post-Training Quantization (PTQ) has emerged as a fundamental technique for deploying large-scale Vision Transformers (ViTs) and Large Language Models (LLMs), as it avoids full model retraining. In the vision domain, methods such as BRECQ, QDrop, and AIQViT employ reconstruction-based calibration to compensate for low-bit perturbations, commonly using gradient-based optimization over blocks or layers \cite{li_brecq_2021,wei_qdrop_2023,jiang2025aiqvit}. In the LLM domain, GPTQ uses approximate second-order information to minimize weight reconstruction error \cite{frantar2022gptq}. AWQ follows a different strategy: rather than backpropagation-based reconstruction, it identifies activation-salient weight channels and applies activation-aware scaling to protect them during weight-only quantization \cite{lin_awq_2024}.

These methods substantially reduce model-storage and memory-bandwidth costs, but their deployment properties differ. Reconstruction-based PTQ can introduce nontrivial offline calibration cost, whereas AWQ avoids backpropagation but still targets weight-only compression. In conventional deployment, the stored integer codes are interpreted through scales, zero-points, codebook decoding, or compensation operations before or around the core arithmetic datapath. Such representations can reduce the memory wall without necessarily removing general multiplication from the dominant computation. GoQuant instead uses an analytical basis construction and PoT-native inner-product operands. It avoids runtime affine reconstruction before the shift-and-add dot-product core, while retaining a small amount of coefficient scaling outside that core.

\subsection{High-Dimensional Geometry and Rotational Quantization}
As traditional 1D scalar quantization approaches practical limits around the 3-bit threshold, recent literature has explored high-dimensional geometric spaces. A prominent trend involves using orthogonal rotations to precondition the weight and activation manifolds prior to discretization. Three notable methodologies represent the state-of-the-art in this rotational paradigm. First, the QuIP family establishes theoretical bounds for sub-3-bit fidelity by utilizing fixed Randomized Hadamard Transforms (RHT) to ensure error incoherence, coupling this with advanced vector projection onto the dense $E_8$ lattice \cite{chee2023quip, tseng2024quip}. Extending this concept, QuaRot applies Hadamard rotations holistically to both weights and activations, smoothing outliers to facilitate uniform low-bit quantization\cite{ashkboos_quarot_2024}. Advancing beyond fixed transforms, SpinQuant introduces parameterized orthogonal matrices, actively rotating the feature spaces via gradient optimization\cite{liu2024spinquant}.

Although mathematically elegant, the paradigm of explicitly rotating data to fit an optimal lattice introduces hardware considerations. While high-performance cloud accelerators may absorb this complexity, applying these dense orthogonal matrices at runtime often requires deep addition trees and non-local memory routing (e.g., butterfly networks). This can exacerbate on-chip wire routing congestion and vector decoding latency. Such architectural complexities present challenges for the minimalist hardware philosophy required for edge computing. This architectural bottleneck motivates the exploration of multiplier-reduced and transform-light networks, which we discuss next.

\subsection{Multiplier-Reduced Architectures and Logarithmic Quantization}
To reduce the reliance on dense MAC units in conventional quantized inference and to avoid the runtime transform overhead introduced by some rotational methods, recent work has increasingly explored hardware-oriented, multiplier-reduced execution paradigms. These approaches commonly employ ternary weights, logarithmic quantization, or bit-level decomposition to replace general multiplications with simpler additions and shifts. Representative examples include BitNet, which trains ternary-weight language models using a specialized pretraining recipe \cite{wang2023bitnet}, and ShiftAddLLM, which converts pre-trained LLM weights into shift-and-add representations through post-training quantization \cite{you_shiftaddllm_2024}. In the vision domain, RepQ-ViT improves hardware compatibility through scale reparameterization, converting channel-wise post-LayerNorm activation quantization and fine-grained logarithmic post-Softmax quantization into more hardware-friendly inference forms \cite{li_repq-vit_2023}.

These approaches expose different trade-offs. BitNet demonstrates the representational potential of ternary-weight models but requires specialized training rather than direct post-training conversion of an existing model. ShiftAddLLM can retrofit pre-trained LLMs, but bit-level decomposition increases the number of additions and may require layer-wise optimization. RepQ-ViT is specifically designed around the activation distributions and operators of Vision Transformers; its scale-reparameterization mechanism should therefore be distinguished from a general per-channel PoT weight format.

Crucially, independent scalar rounding does not explicitly optimize the angular distribution of the PoT representation. GoQuant introduces an alternative view: rather than increasing the density of a scalar codebook or applying an online global rotation, it constructs a local orthogonal residual direction from the primary PoT basis. The proposed hardware-bounded candidate family admits deterministic selection and maps the dominant dot products to shifts and additions. Small coefficient multiplications and compact block-wise exchange metadata remain, so we characterize the complete system as multiplier-reduced rather than universally multiplier-free.

\section{Methodology}

In this section, we present the theoretical formulation and execution pipeline of the GoQuant framework. Departing from conventional one-dimensional scalar quantization, GoQuant interprets ultra-low-bit quantization error as a combination of magnitude and angular deviations. Rather than compensating for these errors solely through scalar rescaling, GoQuant captures the unresolved directional component within a two-dimensional orthogonal subspace constrained by a Power-of-Two (PoT) representation. The framework separates the quantization procedure into two stages: structured discrete basis construction and continuous coefficient optimization.

\subsection{Overall Framework of GoQuant}
Let $\mathbf{w}\in\mathbb{R}^{N}$ denote a weight vector associated with one macro-block, where $N=128$. Traditional PoT quantization approximates $\mathbf{w}$ using a single discrete direction,
\[
\hat{\mathbf{w}}=c_1\mathbf{b}_1,
\]
where $\mathbf{b}_1$ is drawn from a PoT lattice. This one-dimensional representation can exhibit limited angular resolution because the available PoT directions are distributed non-uniformly.

GoQuant addresses this limitation through an orthogonal residual expansion. It constructs a secondary discrete basis $\mathbf{b}_2$ and continuous coefficients $c_1$ and $c_2$ such that
\[
\hat{\mathbf{w}}
=
c_1\mathbf{b}_1+c_2\mathbf{b}_2,
\qquad
\mathbf{b}_1^T\mathbf{b}_2=0.
\]
The first basis captures the dominant PoT approximation, while the second basis is structurally generated to align with the directional residual left by the primary projection.

\subsection{Primary Projection and Orthogonal Residual Extraction}
To effectively utilize the representational capacity of an ultra-low bit hardware budget (e.g., 3-bit or 4-bit), GoQuant transitions from a conventional symmetric projection to a hardware-aware, asymmetric logarithmic lattice. An explicit zero state enables zero-skipping and can be used by hardware control logic to suppress unnecessary switching activity.

For instance, at the 3-bit threshold (i.e., exactly $2^3 = 8$ discrete physical states), instead of relying on external sparsity masks, our quantization function $\mathcal{Q}(\cdot)$ explicitly breaks spatial symmetry by repurposing the smallest positive quantization bin to represent the absolute zero. Consequently, the primary geometric anchor $\mathbf{b}_1$ is constrained to the following asymmetric discrete lattice $\mathcal{L}_{3\text{bit}}$:
\[
\mathcal{L}_{3\text{bit}} = \{-1,-0.5,-0.25,-0.125,0,+0.25,+0.5,+1\}.
\]

Mathematically, let $\mathbf{w}\in\mathbb{R}^{N}$ denote the original weight vector within a 128-dimensional macro-block. We first map it into the lattice domain $[-1,1]$ using the macro-block-wise normalization factor
\[
s_{\mathrm{norm}}=\|\mathbf{w}\|_{\infty}.
\]
The normalized vector is
\[
\mathbf{v}=\frac{\mathbf{w}}{s_{\mathrm{norm}}},
\]
and the primary discrete basis is obtained by round-to-nearest projection:
\[
\mathbf{b}_1=\mathcal{Q}(\mathbf{v}),
\]
where each element is assigned to its nearest valid state in $\mathcal{L}_{3\mathrm{bit}}$.

\begin{figure}[htbp]
    \centering
    \includegraphics[width=0.60\linewidth]{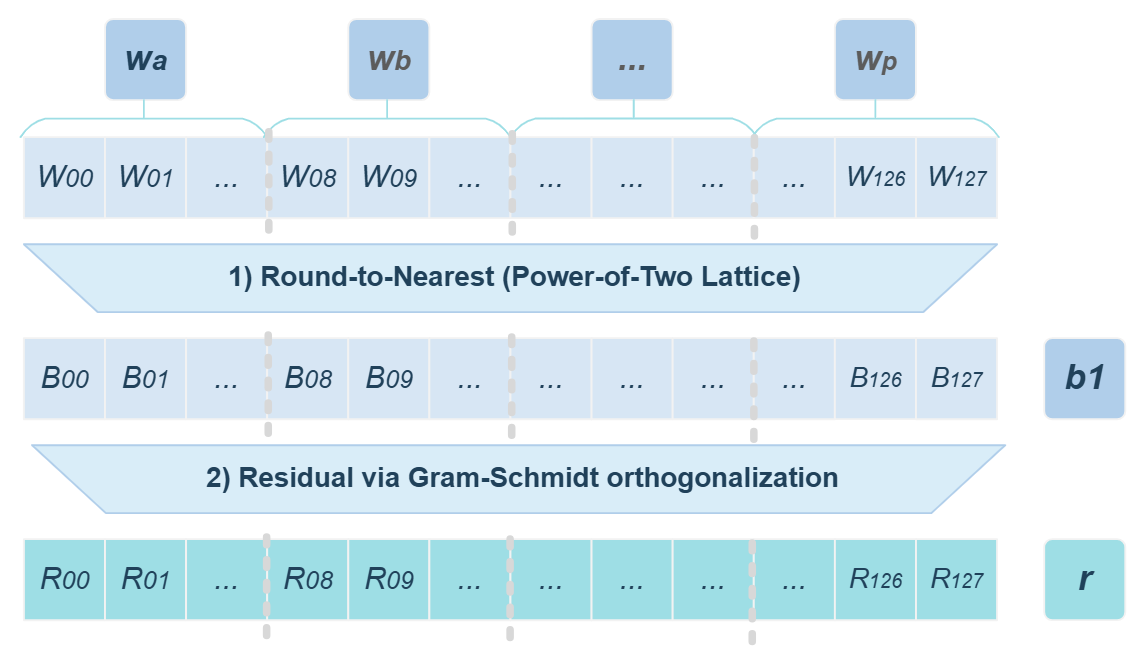}
    \caption{Each 128-dimensional macro-block is partitioned into sixteen 8-dimensional micro-blocks, denoted as $W_a,W_b,\ldots,W_p$. Element-wise round-to-nearest PoT projection constructs the primary basis $\mathbf{b}_1$ under a macro-block normalization factor. The corresponding orthogonal residual $\mathbf{r}_{\perp}$ is then extracted by projection away from the primary direction.}
    \label{fig:s3_reconstr}
\end{figure}

Because this one-dimensional mapping projects a continuous vector onto an exponentially spaced discrete lattice, it can introduce substantial directional error. As illustrated in Fig.~\ref{fig:s3_reconstr}, we isolate the component of the original weight vector that cannot be represented along the primary direction by defining
\begin{equation}
    \mathbf{r}_{\perp}
    =
    \mathbf{w}
    -
    \frac{\langle\mathbf{w},\mathbf{b}_1\rangle}
    {\|\mathbf{b}_1\|_2^2}
    \mathbf{b}_1.
\end{equation}
By construction, $\mathbf{r}_{\perp}^T\mathbf{b}_1=0$. If $\|\mathbf{b}_1\|_2^2=0$, we set $\mathbf{r}_{\perp}=\mathbf{w}$ and bypass the primary projection for that macro-block. The subsequent objective is to construct a secondary discrete basis $\mathbf{b}_2$ that remains orthogonal to $\mathbf{b}_1$ while maximizing its alignment with $\mathbf{r}_{\perp}$.

\subsection{Hierarchical Strided Dual-Exchange and Orthogonal Residual Projection}
A key challenge in extreme low-bit quantization is balancing the memory overhead of scaling factors with the geometric degrees of freedom required for angular fitting. As illustrated in Fig.~\ref{fig:s3_reconstr}, GoQuant employs a \textbf{Hierarchical Block Architecture}. Continuous projection coefficients ($c_1, c_2$) are shared across a coarse macro-block of size $N=128$ to limit memory footprints. Conversely, the generation of the secondary orthogonal basis $\mathbf{b}_2$ is executed at a finer micro-block granularity of $G=8$ (denoted as blocks $a,b,\ldots,p$).

To construct the micro-block components of $\mathbf{b}_2$ that maximize the inner product with the local orthogonal residual $\mathbf{r}_{\perp}$, while maintaining discrete hardware constraints, we propose the \textbf{Strided Dual-Exchange} operator.

\textbf{Theorem 1 (Strided Dual-Exchange Orthogonality).}
Let
\[
\mathbf{v}=[x_0,x_1,\ldots,x_{G-1}]^T
\]
be a $G$-dimensional micro-block sub-vector of the primary basis $\mathbf{b}_1$, where $G=8$. We consider four hardware-supported exchange patterns indexed by
\[
s\in\mathcal{S}=\{1,2,3,4\}.
\]
For each pattern, the exchange partner of index $i$ is defined as
\[
\sigma_s(i)=i\oplus s,
\]
where $\oplus$ denotes bitwise XOR between the 3-bit representations of the indices. Since
\[
\sigma_s(\sigma_s(i))=i
\quad\text{and}\quad
\sigma_s(i)\neq i
\]
for every $s\in\mathcal{S}$, each exchange pattern induces a set of disjoint pairs
\[
\mathcal{P}_s
=
\left\{
(i,\sigma_s(i))
\,\middle|\,
i<\sigma_s(i)
\right\},
\]
such that every coordinate appears in exactly one pair.

For each pair $(i,j)\in\mathcal{P}_s$, let $\eta_{ij}\in\{+1,-1\}$ denote a local sign factor. The corresponding dual-exchanged vector
\[
\mathbf{v}^{*}=[y_0,y_1,\ldots,y_{G-1}]^T
\]
is constructed by applying
\[
\begin{bmatrix}
y_i\\
y_j
\end{bmatrix}
=
\begin{bmatrix}
0 & -\eta_{ij}\\
\eta_{ij} & 0
\end{bmatrix}
\begin{bmatrix}
x_i\\
x_j
\end{bmatrix}
=
\begin{bmatrix}
-\eta_{ij}x_j\\
\eta_{ij}x_i
\end{bmatrix}
\]
to every pair $(i,j)\in\mathcal{P}_s$. Then, for any supported exchange pattern and sign configuration, $\mathbf{v}^{*}$ is orthogonal to $\mathbf{v}$:
\[
\langle\mathbf{v},\mathbf{v}^{*}\rangle=0.
\]

\textit{Proof.}
For each pair $(i,j)\in\mathcal{P}_s$, its contribution to the inner product is
\[
x_i y_i+x_j y_j
=
x_i(-\eta_{ij}x_j)
+
x_j(\eta_{ij}x_i)
=0.
\]
Because the pairs in $\mathcal{P}_s$ are disjoint and collectively cover all coordinates, summing over the complete pairing gives
\[
\langle\mathbf{v},\mathbf{v}^{*}\rangle
=
\sum_{(i,j)\in\mathcal{P}_s}
\left(x_i y_i+x_j y_j\right)
=0.
\]
$\blacksquare$

For $G=8$, the four supported exchange patterns are
\[
\begin{aligned}
\mathcal{P}_1
&=
\{(0,1),(2,3),(4,5),(6,7)\},\\
\mathcal{P}_2
&=
\{(0,2),(1,3),(4,6),(5,7)\},\\
\mathcal{P}_3
&=
\{(0,3),(1,2),(4,7),(5,6)\},\\
\mathcal{P}_4
&=
\{(0,4),(1,5),(2,6),(3,7)\}.
\end{aligned}
\]

\begin{figure}[htbp]
    \centering
    \includegraphics[width=0.60\linewidth]{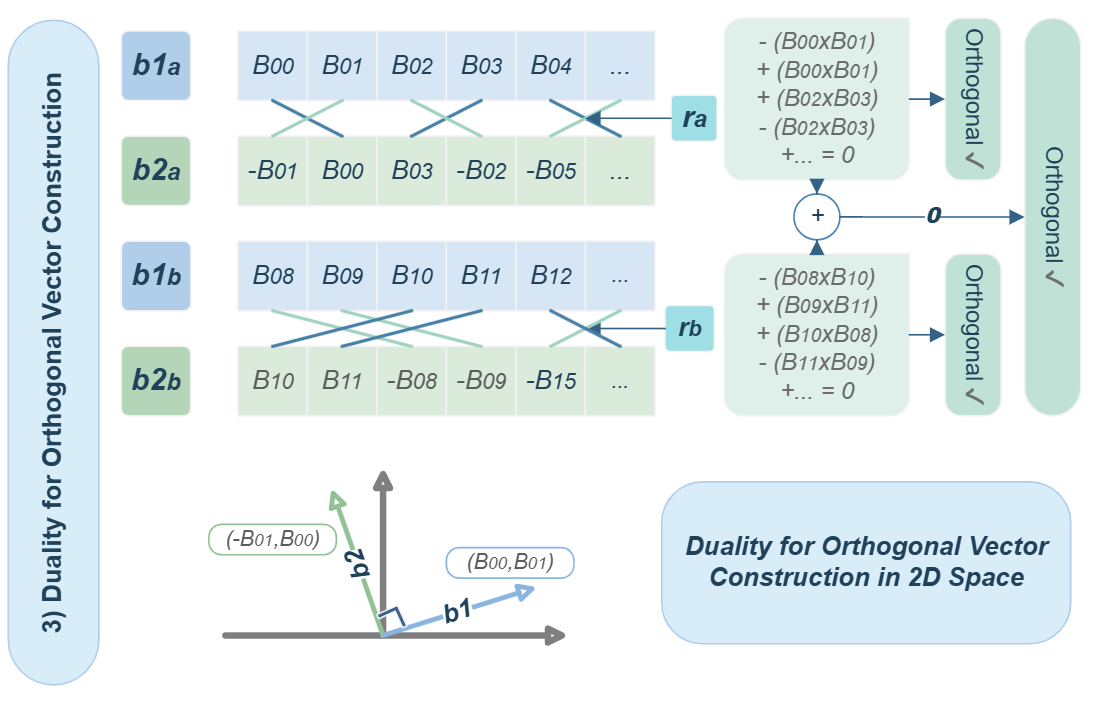}
    \caption{Localized Strided Dual-Exchange within an 8-dimensional micro-block. Each candidate exchange pattern partitions the eight coordinates into four disjoint pairs and applies a signed exchange within every pair. For example, $s=1$ produces adjacent exchanges, whereas $s=2$ produces exchanges separated by two index positions. The resulting secondary basis remains exactly orthogonal to the primary basis. The exchange pattern and sign configuration are determined during quantization, stored as compact micro-block metadata, and used to configure the local routing network during inference without runtime optimization.}
    \label{fig:s3_orth}
\end{figure}

As illustrated in Fig.~\ref{fig:s3_orth}, each signed exchange acts as a structured $90^\circ$ rotation within a local two-dimensional coordinate plane. Orthogonality is therefore guaranteed by construction rather than enforced through an additional numerical constraint or iterative optimization procedure.

The sign inversion in Theorem~1 may algebraically produce a signed value that is not explicitly included in the asymmetric stored codebook $\mathcal{L}_{3\text{bit}}$, such as $+0.125$. This does not require an additional stored magnitude level. After the asymmetric 3-bit code is decoded, the PoT magnitude and execution sign are carried separately in the shift-and-add datapath. The transformation $-\eta_{ij}x_j$ is implemented by routing the decoded magnitude and toggling the execution sign with an XOR operation. The pairwise flip bits are nevertheless model-dependent control metadata and are included in the storage overhead described below.

To construct the macro-block-level secondary basis $\mathbf{b}_2$ that best aligns with the orthogonal residual $\mathbf{r}_{\perp}$, GoQuant employs the following decoupled offline procedure:

\begin{enumerate}
    \item \textbf{Micro-Block Analytical Sign Inference:}
    For a candidate exchange pattern $s$, consider a pair $(i,j)\in\mathcal{P}_s$. Its contribution to the alignment between the secondary basis and the local orthogonal residual is
    \[
    \begin{aligned}
    y_i r_{\perp,i}+y_j r_{\perp,j}
    &=
    (-\eta_{ij}x_j)r_{\perp,i}
    +
    (\eta_{ij}x_i)r_{\perp,j}\\
    &=
    \eta_{ij}
    \left(
    x_i r_{\perp,j}
    -
    x_j r_{\perp,i}
    \right).
    \end{aligned}
    \]
    The optimal sign for this pair is therefore obtained analytically as
    \[
    \eta_{ij}^{*}
    =
    \operatorname{sgn}
    \left(
    x_i r_{\perp,j}
    -
    x_j r_{\perp,i}
    \right),
    \]
    where either sign may be selected when the argument is zero.

    After substituting the optimal pairwise signs, the maximum alignment score associated with exchange pattern $s$ is
    \[
    A_s
    =
    \sum_{(i,j)\in\mathcal{P}_s}
    \left|
    x_i r_{\perp,j}
    -
    x_j r_{\perp,i}
    \right|.
    \]

    \item \textbf{Hardware-Bounded Pattern Selection:}
    For $G=8$, GoQuant evaluates the four supported exchange patterns and selects
    \[
    s^{*}
    =
    \arg\max_{s\in\{1,2,3,4\}} A_s.
    \]
    Each pattern contains four disjoint pairs; therefore, only sixteen pairwise alignment terms are evaluated per micro-block. The resulting configuration is optimal within the proposed four-pattern Strided Dual-Exchange candidate set.

    The restriction to four exchange patterns is also motivated by runtime hardware efficiency. For each output lane, the exchange network selects among at most four candidate partner lanes and can therefore be implemented using a compact 4-to-1 multiplexer. By comparison, a micro-block size of $G=32$ would introduce sixteen candidate exchange partners and require a substantially wider 16-to-1 selection network. Such a network would increase multiplexer depth, cross-lane wiring distance, routing congestion, and critical-path pressure, potentially offsetting the timing benefits of the shift-and-add inner-product datapath.

    \item \textbf{Deployment-Time Metadata and Macro-Block Assembly:}
    The selected exchange pattern $s^{*}$ and its corresponding sign configuration $\{\eta_{ij}^{*}\}$ are determined during quantization and stored as compact micro-block metadata. For $G=8$, the four candidate patterns require a 2-bit pattern index, and the four disjoint pairs require four sign bits. The total structural overhead is therefore 6 bits per 8-weight micro-block, equivalent to 0.75 bit per weight, excluding the PoT codes and the macro-block coefficients. During inference, these stored control bits configure the local 4-to-1 exchange multiplexers and sign XOR gates; no pattern enumeration or sign optimization is performed at runtime.

    The $N/G=16$ configured micro-blocks, denoted as blocks $a,b,\ldots,p$, are concatenated to form the complete 128-dimensional secondary basis $\mathbf{b}_2$.
\end{enumerate}

With the micro-block size fixed at $G=8$, the proposed construction requires only constant local work for each micro-block and scales linearly with the macro-block dimension $N$. More importantly, it converts orthogonal residual projection into a hardware-local combination of bounded lane selection, sign inversion, shifting, and addition. This structured realization avoids high-dimensional iterative optimization and wide global permutation networks while preserving exact orthogonality between the two discrete bases.

By leveraging this hierarchical design and analytical sign formulation, GoQuant selects the best discrete basis within the proposed hardware-bounded candidate family using deterministic $\mathcal{O}(N)$ operations, avoiding the need for exhaustive search in high-dimensional spaces.

\subsection{Joint Scale Optimization: GEO vs. REF Modes}

\begin{figure}[htbp]
    \centering
    \includegraphics[width=0.40\linewidth]{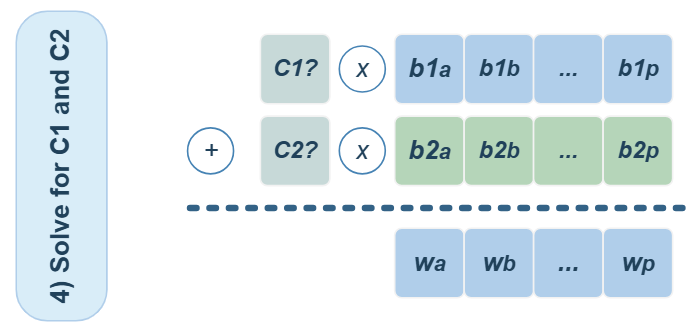}
    \caption{Once the orthogonal bases are anchored, the coefficients $c_1$ and $c_2$ are solved once for each 128-dimensional macro-block and shared across its sixteen micro-blocks, denoted as $a,b,\ldots,p$.}
    \label{fig:s3_solver}
\end{figure}

As illustrated in Fig.~\ref{fig:s3_solver}, once the discrete orthogonal directions $\mathbf{b}_1$ and $\mathbf{b}_2$ are anchored, we solve for the continuous scales $c_1$ and $c_2$, which are shared across a macro-block of size $N=128$ to bound memory overhead. We introduce two specialized solver modes:

\textbf{1. GEO Mode (Geometric Optimization):}
In the absence of calibration data, GEO minimizes the Euclidean reconstruction error directly in the original weight space:
\[
\min_{c_1,c_2}
\left\|
\mathbf{w}-(c_1\mathbf{b}_1+c_2\mathbf{b}_2)
\right\|_2^2.
\]
The optimal coefficients are obtained by solving
\begin{equation}
    \begin{bmatrix}
    \mathbf{b}_1^T\mathbf{b}_1 & \mathbf{b}_1^T\mathbf{b}_2 \\
    \mathbf{b}_2^T\mathbf{b}_1 & \mathbf{b}_2^T\mathbf{b}_2
    \end{bmatrix}
    \begin{bmatrix}c_1\\c_2\end{bmatrix}
    =
    \begin{bmatrix}
    \mathbf{w}^T\mathbf{b}_1\\
    \mathbf{w}^T\mathbf{b}_2
    \end{bmatrix}.
\end{equation}
Because Theorem~1 guarantees $\mathbf{b}_1^T\mathbf{b}_2=0$, the normal equations are diagonal, and the coefficients decouple as
\[
c_1=\frac{\mathbf{w}^T\mathbf{b}_1}{\|\mathbf{b}_1\|_2^2},
\qquad
c_2=\frac{\mathbf{w}^T\mathbf{b}_2}{\|\mathbf{b}_2\|_2^2}.
\]
This closed-form solution enables data-free coefficient estimation without gradient-based optimization.

\textbf{2. REF Mode (Activation-Aware Refinement):}
REF incorporates calibration inputs $\mathbf{X}_q\in\mathbb{R}^{M\times N}$, where $M$ denotes the number of calibration observations and $N=128$ is the macro-block dimension. The output target corresponding to the original weight vector is
\[
\mathbf{Y}=\mathbf{X}_q\mathbf{w},
\]
and the projected basis responses are collected as
\[
\mathbf{A}
=
\left[
\mathbf{X}_q\mathbf{b}_1,\,
\mathbf{X}_q\mathbf{b}_2
\right]\in\mathbb{R}^{M\times2}.
\]
The coefficient vector $\mathbf{c}=[c_1,c_2]^T$ is obtained through ridge regression:
\begin{equation}
    \mathbf{c}^{*}
    =
    \left(
    \mathbf{A}^T\mathbf{A}+\lambda\mathbf{I}
    \right)^{-1}
    \mathbf{A}^T\mathbf{Y},
\end{equation}
where $\lambda=10^{-4}$ is the Tikhonov regularization coefficient. By incorporating the empirical feature covariance induced by $\mathbf{X}_q$, REF prioritizes reconstruction accuracy along directions represented in the calibration data.

It is important to distinguish this coefficient refinement from methods that inject quantized activation noise directly into the design matrix. Such a procedure produces an errors-in-variables setting in which perturbations of $\mathbf{A}$ can bias the least-squares estimate. GoQuant instead constructs the discrete bases independently of activation quantization and uses the calibration features only to solve the two continuous block coefficients.

\subsection{Fully Quantized Inference Dataflow}
For full-layer inference, the macro-block coefficients are collected into coefficient vectors $\mathbf{c}_1$ and $\mathbf{c}_2$, with one scalar coefficient per output macro-block. The dynamic activations $\mathbf{X}$ and these coefficient vectors are quantized to $b_a$-bit and $b_c$-bit integers, denoted as $\tilde{\mathbf{X}}$ and $\tilde{\mathbf{c}}_m$, with scaling factors $s_x$ and $s_{cm}$. The linear-layer computation $\mathbf{Y}=\mathbf{X}\mathbf{W}$ is then expressed as:
\begin{equation}
    \hat{\mathbf{Y}} = (s_x s_{c1}) \left[ (\tilde{\mathbf{X}} \mathbf{B}_1) \text{diag}(\tilde{\mathbf{c}}_1) \right] + (s_x s_{c2}) \left[ (\tilde{\mathbf{X}} \mathbf{B}_2) \text{diag}(\tilde{\mathbf{c}}_2) \right]
\end{equation}

The dominant matrix dot products, $(\tilde{\mathbf{X}}\mathbf{B}_1)$ and $(\tilde{\mathbf{X}}\mathbf{B}_2)$, use PoT operands and are therefore implemented with shifts and additions rather than general multipliers. The subsequent multiplication by the quantized block coefficients $\operatorname{diag}(\tilde{\mathbf{c}}_m)$ remains as a lightweight operation outside the inner-product loop, followed by global rescaling or requantization. GoQuant is therefore multiplier-free in its dominant PoT dot-product core but multiplier-reduced, rather than strictly multiplier-free, at the complete layer level.

\section{Result}

To systematically validate the GoQuant framework, we present an empirical evaluation structured around four core dimensions.

First, we assess the framework's cross-modality applicability by applying reconstruction-free, shift-and-add inner-product constraints to Vision Transformers on the ImageNet-1K benchmark. Second, we examine the relationship between algorithmic fidelity and physical deployment through an Edge NLP algorithm-hardware co-evaluation, integrating LLaMA-2 calibration efficiency with 28\,nm RTL synthesis metrics. Third, we evaluate the architectural design choices via an ablation study on the geometric benefits of dual-basis projection. Finally, we explore the framework's deployment flexibility and resilience to calibration bias through zero-shot reasoning tasks, comparing the GEO and REF paradigms.

\subsection{Main Results on Vision Transformers}

\begin{table*}[t]
\centering
\small
\setlength{\tabcolsep}{4.5pt}
\caption{Top-1 Accuracy (\%) on ImageNet-1K for various Vision Transformers. \textbf{REC} denotes runtime operand reconstruction, not reconstruction-based calibration. A checkmark indicates that the stored low-bit code requires affine decoding (e.g., $\hat{x}=a q+b$), nonlinear codebook interpretation, or an auxiliary compensation step before entering the target core arithmetic datapath. A cross indicates that the encoded values can be consumed directly as native operands by the corresponding arithmetic unit. Common post-accumulation scaling and output requantization are excluded. All GoQuant results use a group size of 128.}
\label{tab:cv_main}
\begin{tabular}{l c c cccccc}
\toprule
\textbf{Method} & \textbf{REC} & \textbf{W/A} & \textbf{DeiT-S} & \textbf{DeiT-B} & \textbf{ViT-S} & \textbf{ViT-B} & \textbf{Swin-S} & \textbf{Swin-B} \\
\midrule
Full-Precision & - & 32/32 & 79.85 & 81.80 & 81.39 & 84.54 & 83.23 & 85.27 \\
\midrule
QDrop\cite{wei_qdrop_2023} & $\checkmark$ & 3/3 & 16.89 & 22.38 & 4.32 & 5.99 & 58.96 & 52.19 \\
BRECQ\cite{li_brecq_2021} & $\checkmark$ & 3/3 & 8.26 & 12.87 & 1.86 & 0.14 & 7.32 & 1.21 \\
AIQViT\cite{jiang2025aiqvit} & $\checkmark$ & 3/3 & \textbf{55.36} & \textbf{66.15} & \textbf{41.32} & \textbf{43.68} & \textbf{71.42} & \textbf{63.01} \\
\cmidrule(lr){4-9}
RepQ-ViT\cite{li_repq-vit_2023} & $\times$ & 3/3 & 4.85 & 7.23 & 0.44 & 0.17 & 1.22 & 4.87 \\
\textbf{Ours} & $\times$ & 3/3 & 22.29& 59.46& 3.19& 50.09& 27.58& 24.97\\
\textbf{Ours} & $\times$ & 3/4 & 69.89& 78.01& 67.00& 79.09& 79.18& 81.72\\
\textbf{Ours} &  $\times$ & 3/6 & 75.99& 80.64& 75.27& 80.51& 81.38& 83.64\\
\midrule
PTQ4ViT\cite{avidan_ptq4vit_2022} & $\checkmark$ & 4/4 & 34.08 & 64.39 & 42.57 & 30.69 & 76.09 & 74.02 \\
APQ-ViT\cite{ding2022towards} & $\checkmark$ & 4/4 & 43.55 & 67.48 & 47.95 & 41.41 & 77.15 & 76.48 \\
QDrop\cite{wei_qdrop_2023} & $\checkmark$ & 4/4 & 35.79 & 65.47 & 17.77 & 21.72 & 78.92 & 80.49 \\
BRECQ\cite{li_brecq_2021} & $\checkmark$ & 4/4 & 54.31 & 62.96 & 63.90 & 61.54 & 76.63 & 74.15 \\
AIQViT\cite{jiang2025aiqvit} & $\checkmark$ & 4/4 & 72.75 & \textbf{79.19} & 70.63 & 74.15 & \textbf{80.93} & 81.22 \\
\cmidrule(lr){4-9}
RepQ-ViT\cite{li_repq-vit_2023} & $\times$ & 4/4 & 69.03 & 75.61 & 65.05 & 68.48 & 79.45 & 78.32 \\
\textbf{Ours} & $\times$ & 4/4 & \textbf{73.53}& 79.10& \textbf{73.02}& \textbf{79.84}& 80.36& 82.68\\
\textbf{Ours} & $\times$ & 4/6 & 78.40& 81.24& 78.98& 81.20& 82.30& 84.48\\
\bottomrule
\end{tabular}
\end{table*}

GoQuant also exhibits favorable adaptability in the sub-4-bit weight regime. Although the fully quantized W3/A3 configuration suffers substantial degradation on several architectures, modestly increasing the activation precision to 4 bits largely restores the lost accuracy. The resulting W3/A4 configuration reaches \textbf{78.01\%} on DeiT-B, \textbf{79.09\%} on ViT-B, and \textbf{81.72\%} on Swin-B, despite using only 3-bit weights. Increasing the activation precision to 6 bits further improves accuracy: W3/A6 achieves \textbf{80.51\%} on ViT-B, 4.03 percentage points below the full-precision baseline of 84.54\%, while reaching 80.64\% on DeiT-B and 83.64\% on Swin-B. Similarly, W4/A6 achieves 81.20\% on ViT-B, corresponding to a 3.34-point gap to full precision, and reaches 84.48\% on Swin-B. These results indicate that GoQuant provides a flexible and hardware-efficient quantization paradigm that generalizes effectively across both language and vision models.

\subsection{Edge NLP Inference and Hardware Co-Evaluation}

\begin{table*}[t]
\centering
\caption{Algorithm--hardware co-evaluation on LLaMA-2 models. Model quality is measured by WikiText-2 perplexity (PPL), and quantization time (QT) is measured on a single workstation. The energy--delay product (EDP) and critical-path delay are obtained by synthesizing the corresponding arithmetic units at a 28\,nm technology node.}
\label{tab:algo_hw_coeval}
\small
\begin{tabular}{c l | c c | c c | c c}
\toprule
& &
\multicolumn{2}{c|}{\textbf{Accuracy (PPL $\downarrow$)}} &
\multicolumn{2}{c|}{\textbf{Quantization Time}} &
\multicolumn{2}{c}{\textbf{Hardware (28\,nm)}} \\
\textbf{W/A} &
\textbf{Method} &
\textbf{L2-7B} &
\textbf{L2-13B} &
\textbf{QT (7B)} &
\textbf{QT (13B)} &
\textbf{EDP (pJ$\cdot$ns) $\downarrow$} &
\textbf{Delay (ns) $\downarrow$} \\
\midrule

\textbf{16/16} & \textbf{Base} &
5.49 & 4.91 &
-- & -- &
-- & -- \\
\midrule

\multirow{2}{*}{4/16}
& AWQ &
\textbf{5.66} & \textbf{5.00} &
15.44 min & 22.80 min &
1.03 & 0.90 \\
& \textbf{GoQuant (Ours)} &
5.83 & 5.10 &
\textbf{4.38 min} & \textbf{10.36 min} &
\textbf{0.80} & \textbf{0.60} \\
\midrule

\multirow{3}{*}{4/8}
& AWQ &
\textbf{5.66} & \textbf{5.00} &
15.78 min & 23.35 min &
0.41 & 0.60 \\
& SpinQuant &
5.70 & \textbf{5.00} &
358.0 min & -- &
-- & -- \\
& \textbf{GoQuant (Ours)} &
5.83 & 5.10 &
\textbf{4.45 min} & \textbf{10.50 min} &
\textbf{0.39} & \textbf{0.35} \\
\midrule

\multirow{2}{*}{3/16}
& AWQ &
6.49 & 5.47 &
15.41 min & 22.68 min &
0.73 & 0.80 \\
& \textbf{GoQuant (Ours)} &
\textbf{6.01} & \textbf{5.22} &
\textbf{4.36 min} & \textbf{10.35 min} &
\textbf{0.49} & \textbf{0.35} \\
\midrule

\multirow{2}{*}{3/8}
& AWQ &
6.50 & 5.47 &
15.75 min & 23.09 min &
0.36 & 0.50 \\
& \textbf{GoQuant (Ours)} &
\textbf{6.02} & \textbf{5.22} &
\textbf{4.45 min} & \textbf{10.50 min} &
\textbf{0.32} & \textbf{0.32} \\
\bottomrule
\end{tabular}
\end{table*}

To evaluate the practical deployment potential of GoQuant, we conduct an algorithm--hardware co-evaluation that jointly considers model quality, quantization time, and RTL-synthesized hardware efficiency. As summarized in Table~\ref{tab:algo_hw_coeval}, GoQuant provides a favorable trade-off between perplexity and deployment cost across different weight and activation precisions.

\begin{figure}[t]
    \centering
    \includegraphics[width=0.35\linewidth]{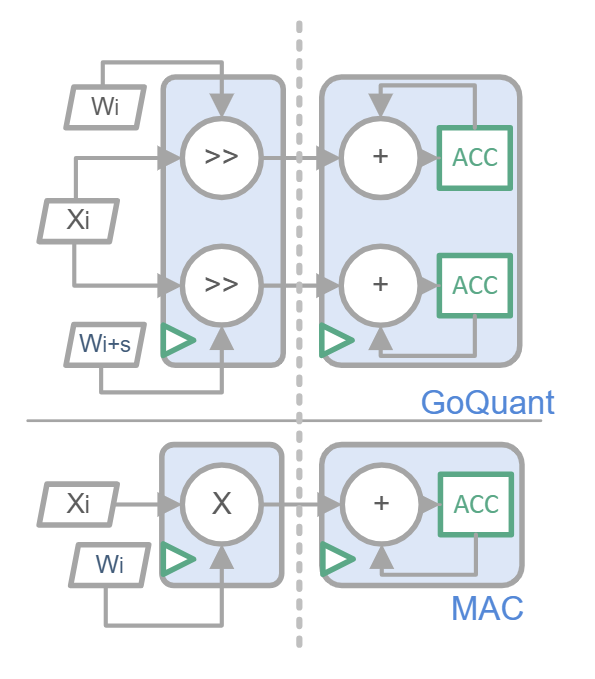}
    \caption{Hardware microarchitecture comparison. The conventional MAC unit (bottom) relies on general multiplication logic, whereas the GoQuant dual-branch inner-product core (top) computes the dominant dot products using parallel bit-shifters (\texttt{>>}) and adders, thereby avoiding general multipliers within the dot-product core. The current RTL synthesis covers only the dual-branch shift-and-add arithmetic core shown here; the exchange-routing network and metadata-control logic are excluded.}
    \label{fig:hardware}
\end{figure}

Figure~\ref{fig:hardware} illustrates the architectural difference between GoQuant and conventional quantization methods such as AWQ and SpinQuant. These baselines generally rely on multiply--accumulate units, whose multipliers and internal adder trees increase the combinational logic depth. In contrast, GoQuant exploits the power-of-two structure of its orthogonal residual representation and replaces multiplication with parallel shifting and addition. This multiplier-free inner-product datapath shortens the critical path and enables higher operating frequencies, while lightweight block-coefficient scaling remains outside the synthesized core comparison.

The synthesized results confirm this hardware advantage. Across all evaluated configurations, GoQuant achieves both a lower EDP and a shorter critical-path delay than AWQ. Under W4/A16, GoQuant reduces the delay from 0.90\,ns to 0.60\,ns and the EDP from 1.03 to 0.80\,pJ$\cdot$ns. Under W3/A16, the delay decreases from 0.80,ns to 0.35\,ns, while the EDP is reduced from 0.73 to 0.49\,pJ$\cdot$ns. Further reductions are observed when the activation precision is lowered to 8 bits. Under W4/A8, GoQuant achieves an EDP of 0.39\,pJ$\cdot$ns and a delay of 0.35\,ns, compared with 0.41\,pJ$\cdot$ns and 0.60\,ns for AWQ. Under W3/A8, GoQuant attains the lowest EDP and delay among all evaluated configurations, reaching 0.32\,pJ$\cdot$ns and 0.32\,ns, respectively. The latter corresponds to a nominal synthesized-unit frequency of approximately 3.13\,GHz.

GoQuant also provides substantial software-level efficiency. It completes quantization in 4.36--4.45 minutes for LLaMA-2-7B and 10.35--10.50 minutes for LLaMA-2-13B. Compared with AWQ, GoQuant is consistently approximately $3.5\times$ faster for the 7B model and approximately $2.2\times$ faster for the 13B model across the evaluated precision settings. For LLaMA-2-7B under W4/A8, GoQuant is also approximately $80\times$ faster than the reported SpinQuant implementation, requiring only 4.45 minutes compared with 358.0 minutes.

In terms of model quality, GoQuant remains competitive with AWQ at 4-bit weight precision, with only a modest perplexity difference. Under both W4/A16 and W4/A8, GoQuant obtains perplexities of 5.83 and 5.10 on LLaMA-2-7B and LLaMA-2-13B, respectively, compared with 5.66 and 5.00 for AWQ. More importantly, GoQuant provides clear improvements in the more challenging 3-bit regime. Under W3/A16, it reduces the LLaMA-2-7B perplexity from 6.49 to 6.01 and the LLaMA-2-13B perplexity from 5.47 to 5.22. Similar improvements are retained under W3/A8, where the corresponding perplexities decrease from 6.50 to 6.02 and from 5.47 to 5.22. Overall, GoQuant combines improved 3-bit model fidelity with substantially faster quantization, lower EDP, and shorter critical paths, making it well suited to efficient low-bit deployment.

\subsection{The Geometric Dividend of Dual-Basis Reconstruction}

\begin{table}[htbp]
  \centering
  \caption{Ablation study on the basis dimension ($K$) for PoT quantization in
  the REF mode. Increasing the basis dimension from $K=1$ to $K=2$ consistently
  improves WikiText perplexity at both 3-bit and 4-bit weight precision, with
  larger gains in the more challenging 3-bit setting.}
  \label{tab:ablation_k_basis}
  \begin{tabular}{c c c c c}
  \toprule
  \textbf{Topology} & \textbf{W/A} & \textbf{Dim ($K$)}
  & \textbf{L2-7B $\downarrow$}
  & \textbf{L2-13B $\downarrow$} \\
  \midrule
  \multirow{4}{*}{\textbf{PoT}}
  & \multirow{2}{*}{3/16}
  & $K=1$ & 6.72 & 5.57 \\
  & & \textbf{$K=2$} & \textbf{6.01} & \textbf{5.22} \\
  \cmidrule{2-5}
  & \multirow{2}{*}{4/16}
  & $K=1$ & 5.97 & 5.24 \\
  & & \textbf{$K=2$} & \textbf{5.83} & \textbf{5.10} \\
  \bottomrule
  \end{tabular}
\end{table}

To validate the proposed dual-basis design, we study the effect of the basis dimension ($K$) for PoT quantization in the REF mode. As shown in Table~\ref{tab:ablation_k_basis}, increasing the basis dimension from $K=1$ to $K=2$ consistently improves WikiText perplexity at both 3-bit and 4-bit weight precision, while the magnitude of the improvement depends on the severity of the quantization constraint.

The dual-basis design provides particularly clear benefits in the more challenging W3/A16 setting. For LLaMA-2-7B and LLaMA-2-13B, increasing $K$ from 1 to 2 reduces perplexity from 6.72 to 6.01 and from 5.57 to 5.22, corresponding to absolute improvements of 0.71 and 0.35, respectively. The dual-basis representation also remains beneficial under W4/A16, reducing perplexity from 5.97 to 5.83 for LLaMA-2-7B and from 5.24 to 5.10 for LLaMA-2-13B, corresponding to improvements of 0.14 in both cases. These results indicate that the additional orthogonal basis consistently improves the representational fidelity of PoT quantization, with a more pronounced contribution when the primary PoT lattice is more severely constrained.

\subsection{Zero-Shot Reasoning and The Dual-Track Deployment Paradigm}

\begin{table}[t]
\centering
\caption{Performance and deployment efficiency of the GoQuant framework on LLaMA-2 models compared to the FP16 baseline. All experiments were conducted on a single NVIDIA RTX PRO6000 (96GB) GPU. GEO and REF denote the data-free mode and activation-refined mode, respectively.}
\label{tab:main_results}
\small
\setlength{\tabcolsep}{4.5pt}
\begin{tabular}{l c l c c c c c}
\toprule
\textbf{Model} & \textbf{W/A} & \textbf{Mode} & \textbf{Time} & \textbf{WK$\downarrow$} & \textbf{HS$\uparrow$} & \textbf{ARC-C$\uparrow$} & \textbf{WG$\uparrow$} \\
\midrule
\multirow{5}{*}{L2-7B}
 & 16/16 & FP16 & -- & 5.49 & 57.14\% & 43.34\% & 69.06\% \\
\cmidrule{2-8}
 & 4/16 & Go-GEO &  0.47 min & 6.01& \textbf{55.45\%}& 41.47\%& 68.59\%\\
 & 4/16 & Go-REF & 4.38 min& \textbf{5.83} & 55.30\%& \textbf{41.81\%}& \textbf{69.38\%}\\
\cmidrule{2-8}
 & 3/16 & Go-GEO & 0.47 min&  7.90& 50.40\%& 39.25\%& 66.77\%\\
 & 3/16 & Go-REF & 4.36 min& \textbf{6.01}& \textbf{54.80\%}& \textbf{40.96\%}& \textbf{68.98\%}\\
\midrule
\multirow{5}{*}{L2-13B}
 & 16/16 & FP16 & -- & 4.91 & 60.07\% & 48.46\% & 72.30\% \\
\cmidrule{2-8}
 & 4/16 & Go-GEO & 0.96 min & 5.21 & 58.77\% & \textbf{46.84\%} & \textbf{71.98\%} \\
 & 4/16 & Go-REF & 10.52 min & \textbf{5.10} & \textbf{59.12\%} & 46.16\% & 71.43\% \\
\cmidrule{2-8}
 & 3/16 & Go-GEO & 0.95 min & 6.91 & 54.56\% & 41.55\% & 70.40\% \\
 & 3/16 & Go-REF & 10.54 min & \textbf{5.22} & \textbf{58.50\%} & \textbf{43.86\%} & \textbf{72.06\%} \\
\bottomrule
\end{tabular}
\end{table}

Table~\ref{tab:main_results} evaluates GoQuant on LLaMA-2 under low-bit weight quantization, reporting WikiText-2 (WK) perplexity and zero-shot accuracy on HellaSwag (HS), ARC-Challenge (ARC-C), and WinoGrande (WG). A key advantage of GoQuant is its calibration efficiency. The data-free Geometric-Only (GEO) mode completes full-model quantization in only 0.47 minutes for LLaMA-2-7B and approximately 0.95--0.96 minutes for LLaMA-2-13B. Even the Activation-Refined (REF) mode requires only 4.36--4.38 minutes for the 7B model and approximately 10.52--10.54 minutes for the 13B model on a single NVIDIA RTX PRO6000 GPU. These results demonstrate that GoQuant can instantiate practical low-bit models within minutes without relying on computationally expensive iterative search.

The GEO mode provides a competitive data-free solution, particularly under W4/A16 quantization. For LLaMA-2-7B, Go-GEO achieves a WK perplexity of 6.01 and retains 68.59\% accuracy on WinoGrande. For LLaMA-2-13B, it reaches a perplexity of 5.21, an ARC-C accuracy of 46.84\%, and a WinoGrande accuracy of 71.98\%. The REF mode further improves quantization fidelity by incorporating activation statistics from a small calibration set. Its benefits are especially pronounced in the more challenging W3/A16 setting. For LLaMA-2-7B, REF reduces WK perplexity from 7.90 to 6.01 and improves HellaSwag, ARC-C, and WinoGrande accuracy from 50.40\%, 39.25\%, and 66.77\% to 54.80\%, 40.96\%, and 68.98\%, respectively. Similarly, for LLaMA-2-13B, REF reduces perplexity from 6.91 to 5.22, while increasing HellaSwag accuracy from 54.56\% to 58.50\% and WinoGrande accuracy from 70.40\% to 72.06\%.

At W4/A16, the REF models remain close to their FP16 counterparts. The LLaMA-2-13B REF model achieves a WK perplexity of 5.10 and a HellaSwag accuracy of 59.12\%, compared with 4.91 and 60.07\% for the FP16 baseline, respectively. The LLaMA-2-7B REF model additionally reaches 69.38\% on WinoGrande, slightly surpassing the FP16 baseline of 69.06\%. These results indicate that a shift-and-add PoT inner-product core, combined with lightweight coefficient scaling, can preserve strong representational fidelity at scale.

The comparison between GEO and REF also reveals a task-dependent trade-off. For LLaMA-2-13B at W4/A16, REF improves WikiText-2 perplexity from 5.21 to 5.10 and HellaSwag accuracy from 58.77\% to 59.12\%. In contrast, GEO performs better on ARC-C and WinoGrande, achieving 46.84\% and 71.98\%, compared with 46.16\% and 71.43\% for REF. This behavior suggests that activation-based refinement primarily improves calibration-domain reconstruction but does not necessarily produce uniform gains across all downstream tasks. GEO, by contrast, avoids dependence on calibration data and therefore provides a fast and robust data-free baseline. Together, the two modes allow GoQuant to flexibly balance calibration cost, quantization fidelity, and cross-task generalization according to deployment requirements.

\section{Conclusion}

In this paper, we propose the Geometric Orthogonal Residual Projection Quantization (GoQuant) framework, an algorithm--hardware co-design approach for ultra-low-bit model quantization. Departing from conventional scalar-lattice fitting, GoQuant represents quantization as a geometric projection onto structured orthogonal PoT bases. This formulation enables the dominant dot-product computation in both LLaMA-2 and Vision Transformers to be mapped onto a shift-and-add inner-product datapath without runtime affine reconstruction of the stored PoT operands.

Our empirical evaluation demonstrates a favorable balance among model fidelity, calibration efficiency, and deployment cost. On Vision Transformers, GoQuant remains effective across multiple architectures and supports flexible weight--activation precision configurations. On LLaMA-2, GEO completes full-model quantization in less than one minute, while REF requires approximately 4.4 minutes for the 7B model and 10.4--10.5 minutes for the 13B model. The two modes provide a practical choice between data-free deployment and activation-aware coefficient refinement.

GoQuant is particularly effective in the challenging 3-bit regime. Compared with AWQ, it improves WikiText-2 perplexity under both W3/A16 and W3/A8 while reducing quantization time by approximately $3.5\times$ for LLaMA-2-7B and $2.2\times$ for LLaMA-2-13B. At the hardware level, the PoT inner-product core replaces dense multiplier trees with parallel shifts and additions. Standard-cell synthesis at a 28\,nm technology node shows lower energy--delay products and shorter critical-path delays than the corresponding AWQ arithmetic units across all evaluated configurations. Under W3/A8, the minimum critical-path delay is 0.32\,ns, corresponding to a nominal synthesized-unit frequency of approximately 3.13\,GHz.

The complete layer remains multiplier-reduced rather than universally multiplier-free because lightweight block-coefficient scaling is retained outside the dominant dot-product loop. In addition, the selected stride and pairwise flip configuration is stored as compact micro-block metadata, amounting to 0.75 bit per weight for $G=8$. Overall, GoQuant demonstrates that a geometric orthogonal-residual view can substantially improve the effective resolution of an ultra-low-bit PoT representation while preserving a timing-oriented hardware structure for resource-constrained edge deployment.

\bibliographystyle{unsrt}
\bibliography{references}

\end{document}